\documentclass{article}
\usepackage{ifpdf}
\usepackage{times}
\usepackage{epsfig}
\usepackage{graphicx} 
\usepackage{subfigure}

\usepackage{natbib}

\usepackage{algorithm}
\usepackage{algorithmic}
\usepackage{enumitem}  

\usepackage{hyperref}

\newtheorem{defn}{Definition}[section]

\usepackage[accepted]{icml2016}

\usepackage{amsmath,amssymb, gensymb}
\usepackage{multirow}
\date{}
\begin{document}

\title{An end-to-end convolutional selective autoencoder\\approach to Soybean Cyst Nematode eggs detection}

\author{
Adedotun Akintayo\\
{\tt\small akintayo@iastate.edu}
\and
Nigel Lee\\
{\tt\small nigel92@iastate.edu}
\and
Vikas Chawla\\
{\tt\small vchawla@iastate.edu}
\and
Mark Mullaney\\
{\tt\small mmullane@iastate.edu}
\and
Christopher Marett\\
{\tt\small cmarett@iastate.edu}
\and
Asheesh Singh\\
{\tt\small singhak@iastate.edu}
\and
Arti Singh\\
{\tt\small arti@iastate.edu}
\and
Greg Tylka\\
{\tt\small gltylka@iastate.edu}
\and
Baskar Ganapathysubramaniam\\
{\tt\small baskarg@iastate.edu}
\and
Soumik Sarkar\\
{\tt\small soumiks@iastate.edu}
}


\maketitle
\vskip 0.3in

\begin{abstract}
This paper proposes a novel selective autoencoder approach within the framework of deep convolutional networks. The crux of the idea is to train a deep convolutional autoencoder to suppress undesired parts of an image frame while allowing the desired parts resulting in efficient object detection. The efficacy of the framework is demonstrated on a critical plant science problem. In the United States, approximately $\$1$ billion is lost per annum due to a nematode infection on soybean plants. Currently, plant-pathologists rely on labor-intensive and time-consuming identification of Soybean Cyst Nematode (SCN) eggs in soil samples via manual microscopy. The proposed framework attempts to significantly expedite the process by using a series of manually labeled microscopic images for training followed by an automated high-throughput egg detection. The problem is particularly difficult due to the presence of a large population of non-egg particles (disturbances) in the image frames that are very similar to SCN eggs in shape, pose and illumination. Therefore, the selective autoencoder is trained to learn unique features related to the invariant shapes and sizes of the SCN eggs without hand-crafting following which a composite non-maximum suppression and differencing is applied at the post-processing stage.

\end{abstract}

\section{Introduction}
\label{sec:intro}
With the versatility of machine learning tools expanding to various scopes
~\cite{YB08}, neural networks are particularly becoming popular for a wide variety of applications. For instance, convolutional neural network-based applications include Graph Transformer Networks, GTN for rapid, online recognition of handwriting~\cite{LBBH98}, natural language processing~\cite{CW08}, large vocabulary continuous speech recognition ~\cite{SPKL15} and avatar CAPTCHA machine image recognition~\cite{BC12} by training machines to distinguish between human faces and computer generated faces. In the area of object detection, while dimension reduction techniques such as principal component analysis (PCA)~\cite{BF07}, Independent Component Analysis and Linear Discriminant Analysis, LDA ~\cite{MK01} have been widely used, Support vector machines, SVM~\cite{B98} are quite successful for similar applications as well. They provide competitive prior knowledge-free mapping of complex planes in images to their high dimensional space. Part-based discriminative model~\cite{FGMR10}, and exemplar-SVMs~\cite{MGE11} have been tested to tackle the Pascal visual object classes (VOC) challenges and robotics applications. However, the presence of a large amount of data (microscopic images of SCN eggs with significant intra-class varieties and non-uniformly illumination), the intent of avoiding hand-crafted features and availability of powerful computing infrastructure made deep neural networks a suitable candidate for the current application. From a model architecture perspective, multi-scale convolutional networks for scene labeling~\cite{FCNL13} and pool-based, pylon model segmentation from trees~\cite{LVZ11} having similarities in multi-scale approach on superpixels. However, the class of objects in this case have high similarities than the results obtained with direct application of either models. Application of multiscale architectures in this case would, in principle be detrimental to the performance of an autoencoder that is trained to be selective. 

\begin{figure}[!htb]
\vskip -0.05in
\begin{center}
\centerline{\includegraphics[width=\columnwidth]{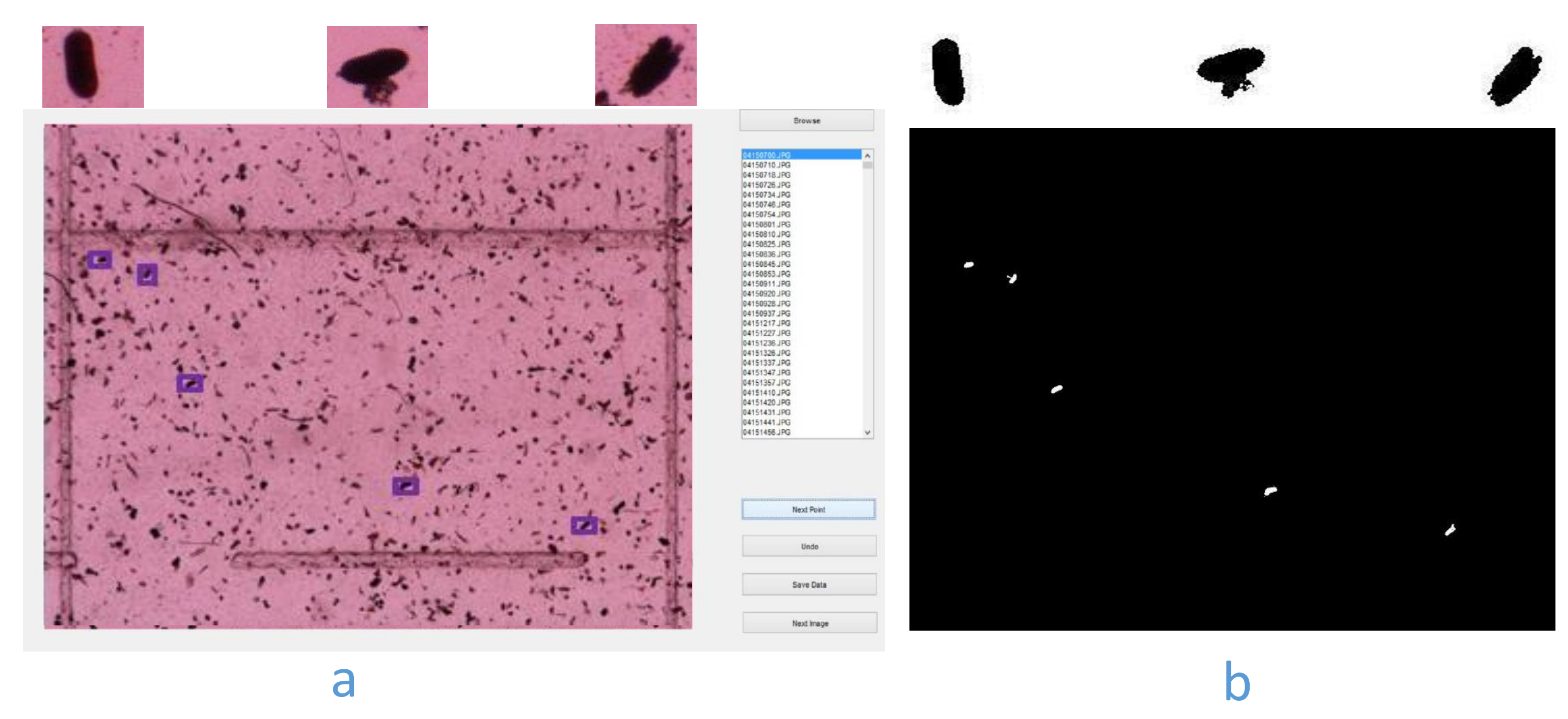}}
\caption{Plate a). shows example of frames with eggs in purple boxes and unwanted particles on a MATLAB-based GUI for tool-labeling of plant-pathologists-identified eggs and Plate b). shows the ground truth labels derived}
\label{fig:example}
\end{center}
\vskip -0.22in
\end{figure}

Soybean Cyst Nematodes are unwanted microorganism that are known to compete with the roots of soybean plants for available nutrients causing stuntedness, limiting nodulation of nitrogen fixations and therefore large yield loss of between $30 - 100 \%$ ~\cite{J13, ISU08}. The Cysts are formed by dead female worms which, prior to dying already secreted the eggs and still provide suitable condition for their continuous development. The challenge therefore is to isolate eggs from many other particles in a soil sample. The current practice is to manually identify and count such eggs under a microscope which is an extremely tedious and time-consuming effort while being significantly error-prone. Unfortunately, numerous computer vision-based automation attempts have failed as the problem is extremely nontrivial due the rarity of SCN egg present on a typical microscopic image frame and they have great similarities with various non-egg objects on those frames. Thus, isolating the rarely present SCN eggs from other undesired non-eggs particles on microscopic image frames (as shown in fig.~\ref{fig:example}) is a complex object detection problem that have enormous plant science implications.
In this context, this work attempts to develop an efficient, high throughput, end-to-end SCN egg detection solution using a novel convolutional selective autoencoder approach.
The primary contributions of this work are summarized as follows:
\begin {itemize}[noitemsep,nolistsep]
\item novel selective autoencoder approach - to train a deep convolutional autoencoder to suppress undesired parts of an image frame while allowing the desired parts resulting in efficient object detection.\\
\item demonstrating the efficacy of the proposed method on a new impactful plant science application involving rare object detection in a microscopic image frame cluttered with disturbances having great similarities with the objects of interest (typically $< 5\%$ SCN eggs among all objects).\\
\item differencing enhanced non-maximum post-processing for improving detection performance.
\end{itemize} 
\subsection{Paper layout}
The paper was introduced in this section with some motivations to the importance of the research. Section~\ref{sec:priorwork} describes the prior work for the platform and an inspiration for improvement. Section~\ref{sec:method} is devoted to formulation and description of the composite architecture. In section~\ref{sec:implementation}, dataset generation is discussed and algorithm's implementation are shown. Results shown are analysed and discussed in section~\ref{sec:results}. The summary, conclusions and future directions are provided in the concluding section.
\section{Background and motivation}\label{sec:priorwork}
\subsection{Convolutional networks}
Convolutional networks are discriminative models that rely primarily on local neighborhood matching for data dimension reduction using nonlinear mapping (i.e. sigmoid, softmax, hyperbolic tangent). Each unit of the feature maps has common shared weights or kernels for efficient training in relatively - compared to fully connected layers - lower trainable parameters, added to an additive bias on which is squashed. Feature extraction and classifier learning are the two main functions of these networks~\cite{LBBH98}. However, to learn the most expressive features, we have to determine the invariance rich codes embedded in the raw data and then follow with a fully connected layer to reduce further the dimensionality of the data and map the most important codes to a low dimension of the examples. Many image processing and complex simulations depend on the invariance property of the convolution neural network stated in ~\cite{LB95} to prevent overfitting by learning expressive codes.
The feature maps are able to preserve local neighborhood patterns for each receptive field as with over-completeness dictionary in ~\cite{AEB06}. The fully connected layers tend to complement the learned features by propagating only the highly active weights and serving as the classifier. A full and detailed review may be found in ~\cite{LBBH98} where the authors note the advantage of local correlation enforcing convolution before spatio-temporal recognition. For efficient learning purposes, convolutional networks are able to utilize distributed map-reduce frameworks~\cite{FM04} as well as GPU computing.

\subsection{Selective autoencoders}
\begin{figure}[!htb]
\vskip -0.1in
\begin{center}
\centerline{\includegraphics[width=\columnwidth]{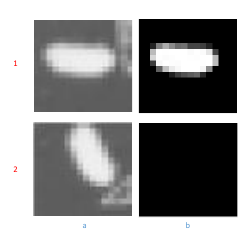}}
\caption{Plate 1a). shows an example of a fully centered egg, 1b). shows its training label, 2a). shows a rotated version of the same egg and 2b). shows how it is patch-blocked in labeling}\vspace{-15pt}
\label{fig:pbs}
\end{center}
\vskip -0.3in
\end{figure}

Deep autoencoders typically extract hierarchical features leading to a relatively small code layer to capture succinct information regarding an input image such that it can be reliably reconstructed through the decoding layers. Among various uses, denoising autoencoders~\cite{VLB08} are particularly interesting as they help denoise input images by reconstructing cleaner versions of them. With a similar motivation, we train a deep convolutional autoencoder to suppress undesired parts (non-egg objects) of an image frame while allowing the desired parts (egg objects) resulting in efficient object detection. In this process, an image frame is divided into many smaller patches, where a patch is labeled as an egg patch
when a full SCN egg is enclosed and centered in a patch while a negative example is when there is no egg present in a patch or there is an egg that is neither full nor centered as shown in fig.~\ref{fig:pbs}. Similar idea can be found in~\cite{MGE11} where an exemplar-based SVM is applied on a neuro-psychological problem. Other similar formulations in~\cite{KRL91, MBCD92} are sometimes called centering. Also, in~\cite{LBBH98} a segmentation graph as prior knowledge was considered with the aim of learning better features.
\section{Algorithm description} \label{sec:method}
\begin{figure*}[!htb]
\centering
 \includegraphics[width=1.0\textwidth,trim={0 0 10 0}]{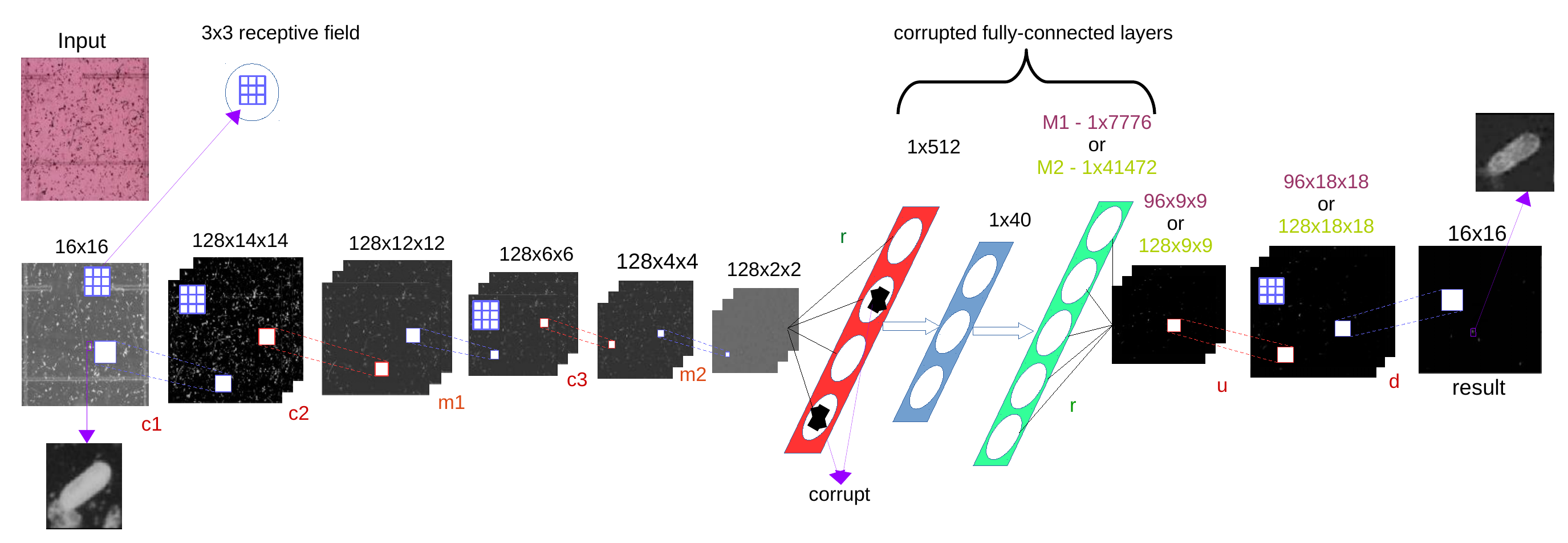}
\caption{convolutional autoencoder architecture for two alternative model structures, Model1 - M1(compressed decoder) and Model2 - M2(uncompressed decoder). Letters are used to describe the layer types, c- convolution, m-maxpooling, r-reshape, u-unpooling and d-deconvolution layers respectively and the digits denote the numbered position of that layer among other layers of same alphabet}
\label{fig:architecture}
\vskip -0.20in
\end{figure*}

Based on convolutional network's (convnet) performances on several tasks reviewed, an end-to-end convolutional autoencoder (as shown in fig.~\ref{fig:architecture}) is designed for the current problem.
Given an $M \times N$ dimensional image frame, P number of patches of dimension, $m \times  n$ were extracted from the image ensuring adequate localisation of algorithm on frames. While an original patch is denoted by $X^i$, the corresponding label patch is denoted by $Y^i$ for $i = 1, 2, \cdots, P$. The constituent layers in the model learning steps from data are outlined as follows:\\
\textbf{Preprocessing \& patch labels}: Data pairs $\{(X^1, Y^1), \cdots, (X^P, Y^P) \}$ are globally normalized together. Furthermore, in order to reduce the probability of false alarms while enhancing egg similar shape, size and pose only, those patches are blocked (considered negative example) where the non-egg varieties are extremely similar in shape or some eggs are partly visible. \\
\textbf{Convnet layers}: At each convolution or deconvolution layer, a chosen $(c \times c)$ filter size is convolved with the patches to learn a $z_o-$dimensional feature map from which joint weight over the $z_i-$dimensional feature maps that are useful for enforcing local correlation is learnt to characterize all maps as follows,
\begin{equation}
\hat{Y}_{z_o(m-c+1)(n-c+1)} = C[X_{z_i m n} \star W_{z_i c c} + b_c ]
\end{equation}
where $C$ is the squashing function, rectified linear unit used and $\star$ is a convolution operator of the joint weights, $W_{z_i c c} $, $b_c$ the biases and input from previous layer, $X_{z_i m n}$.\\
To enhance the invariance further, pooling is done to select representative features in a local neighborhood. It ensures that the neurons activation in a locality do not all favor high entropy in which case information gets diffused. In this formulation, maxpooling~\cite{SMB10} was selected as a representative for a $p \times p$ neighborhood.
\begin{equation}
\hat{Y}_{z_i k l} = \underset{\substack{i \in I; \ i \rightarrow i+p \\ j \in J; \ j \rightarrow j + p }} {\max} (\hat{Y}_{z_i ij})
\end{equation}
where $z_i$ is the number of receptive fields of the input feature maps, $\hat{Y}_{z_i k l}$ is the pooled feature map, $\hat{Y}_{z_i ij}$ is the input from a previous layer, and $I = \{1+(k-1)(p+1), \cdots \}$,  $J = \{1+(l-1)(p+1), \cdots \}$ where $i = 1,2,...,h$ and $j=1,2,...,v$ and $h,v$ denote the horizontal and vertical input dimensions respectively.\\
\textbf{Corruption-induced fully connected layers}:
After flattening features maps from the convolution and subsampling layers, the input features to this layer, say $Y$, are corrupted with random Gaussian noise to produce $\hat{Y}$. This was done to utilize some already experimented benefits of the denoising autoencoder architecture proposed by ~\cite{VLB08}.
\begin{equation}
\hat{Y}_e= E[W_e \hat{Y}+ b_e ]
\end{equation}
and the decoder is given by
\begin{equation}
\hat{Y}_d= D[W_d \hat{Y}_e+ b_d ]
\end{equation}
where $D$ and $E$ stands for the decoder and encoder function respectively, and are also rectified linear units, ReLU. The biases are subscripted, $b_e$ and $b_d$ with alphabet for each layer respectively while the  weights, $W_e$ and $W_d$ are not necessarily tied.
In the foregoing, the nonlinearity function ~\cite{EY14}, rectified linear unit is given by,
\begin{equation}
ReLU(f) = max(0,f)
\end{equation}
It intuitively has the advantage of maximizing the likelihoods whenever there is an egg patch. \\
\textbf{Unpool}:
In this layer, a reversal of the pooled dimension is done by stretching and widening ~\cite{SJ15} the identified features from the filters of the previous layer. It is also an upscaling of the activation around the symmetry lines of each feature map which would then be optimized by the back-propagation algorithm.\\
\textbf{Error minimization}:
The training process includes a regularization function as in the ~\cite{LBBH98} without which the error profile would not generally be monotonically decreasing. The nesterov momentum-based ~\cite{SMDH13} stochastic gradient descent was used for improved results when compared to other loss functions: adaptive subgradient, ADAGRAD ~\cite{DHS11}, Adaptive learning rate method, ADADELTA~\cite{MZ12}, for the reconstruction error updates.  Given the reconstructed output, $\hat{Y}_{mn}$ and the labels, ${Y}_{mn}$. Let $\theta = \{\textbf{W}, \textbf{b}\}$ be the set of weights and biases respectively for all layers, the loss function, $L(\theta)$ which is minimized at each time steps, like other layers during the back-propagation algorithm. It is expressed as,
\begin{equation}
L(\theta) = L_{train}(\theta) + \sigma R(W)
\end{equation} \label{equ:tot_loss}
where $\sigma$ is a parameter controlling the regularization function, $R(W)$;
\begin{equation}
R(W) = (\sum_l \sum_{W_{dim}} W_{l}^2)^{1/2} +\sum_l \sum_{W_{dim}} |W_{l}|
\end{equation}
where $l$ represents layer and $W_{dim}$ represents the dimension of the weight at each layer.  Even though~\cite{YB08} points that SGD with early stopping is equivalent to an $\ell^2-$ regularization. The mean square error training loss is given by,
\begin{equation}
L_{train}(\theta) = \frac{1}{m \times n}\sum_{i=1}^m \sum_{j=1}^n (Y_{ij} - \hat{Y}_{ij})^2
\end{equation}
Then the weights are updated for each time step of the stochastic gradient descent is updated as explained in \cite{LBBH98} to be
\begin{equation}
W_t = W_{t-1} - \alpha \frac{\partial L(W)}{\partial W}
\end{equation}
where $\alpha$ is the learning rate equivalent of step size in optimisation problems.\\
More details of the background can be found in ~\cite{MMCS11} while the so far described and those in section~\ref{sec:implementation} are the more important aspects and improvements made.
\section{Dataset and implementation}
\label{sec:implementation}
\textbf{Dataset generation}:
The dataset is typically generated using a 1-inch-diameter soil probe to collect soil. Soil was collected during Fall 2015 from random placement of soil probe within several farms in the state of Iowa exhibiting different levels of SCN infestation. Each soil sample is mixed together in a bag and washed with water. Purple dye is then applied to the soil and the soil is sonicated to break apart the sac releasing the SCN eggs. A small sample is put on a cover slip and images of the sample were taken using a camera through a microscope.
About a thousand images were collected using this protocol. These images were then labeled by trained plant-pathologists. Labeling consisted of carefully screening each image and identifying the location of every SCN egg present in that image. To enable efficient labeling, a Matlab based app with GUI shown in fig~\ref{fig:example} was created to simplify identification and marking of the SCN eggs location in the image. The app design included a user-friendly way of selecting images, zoom functions, drawing a rectangle region of interest over the eggs, saving the location and skipping an image if no SCN eggs were found. The app was deployed on a touch screen enabled device like the Microsoft Surface Pro, allowing the plant-pathologists who detects  the eggs physically to just use their fingertips for rapid labeling. The bounding box of every SCN egg in the 644 images was extracted and stored. \\
\textbf{Training set}:
This is divided mainly into the cropped and labeled training image which were $11989$ each with frames of ($64 \times 64$), the labeled sets which were $644$ in number each with dimensions  $(2560 \times 1920 )$. However, only $634$ of the latter was used for training while $10$ was randomly left out to test the model.
For uniformity, sets were resized down to the patch size, $(m \times n)$ = $(16 \times 16)$ while the unsegmented frames $(640 \times 480)$ were patched and vectorized to the same size patches and both are concatenated.  The total set available to train the model is shown on  table~\ref{train-table} after the transformation which includes rotation of each egg between $0 -180\degree$ to cover the input space of its variety in orientation as this helps reduce parameters to learn. Training dataset was made up of $80\%$ for training and $20\%$ for validation.\\
\begin{table}[t]
\caption{Training set breakdown: cropped - S, translated - T, rotated - R and labeled - L}  
\label{train-table}
\begin{center}
\begin{small}
\begin{sc}
\begin{tabular}{lcccr}
\hline
\abovespace\belowspace
set type & original dimension & final dimension\\
\hline
\abovespace
s,t,r \& l   & $45432 \times 10$ rotations&$454320\times 16 \times 16 $ \\
s, r \& l  & $2524 \times 10$ rotations& $25240 \times 16 \times 16$\\
l only  & $634 \times 480 \times 640$ & $760800 \times 16 \times 16$\\
\belowspace
  Total &   & $1240360 \times 16 \times 16$ \\
\hline
\end{tabular}
\end{sc}
\end{small}
\end{center}
\end{table}\\
\textbf{Training process}:
Training the problem required that the learning rate was kept low at $0.0001$ with the momentum rate of $0.975$ to prevent oscillations about the minima. The trade-off was that training for several more epochs, to about $100$ was used for this model. As stated earlier, $\ell^1$ and $\ell^2-$regularization parameters of $0.0001$ each were added to widen the parameter search space for locating the minima since that helps to minimize the difference between the test and training.\\
The training was done on GPU Titan Black with 2880 CUDA cores, 6GB memory, in the theano~\cite{BBBLPDTFB10}, lasagne and nolearn wrappers ~\cite{MT15} of python based on improvements described in Section~\ref{sec:method}. Lasagne had the layer details, nonlinearity types, objective function, theano extension and many more built into it. Nolearn on the other hand was a coordinating library for the implementation of the layers in lasagne including the visualization aspects. In the training section, a ($c \times c$) = ($3 \times 3$) filter size and a non-overlapping ($p \times p$) = ($2 \times 2$) were found to be experimentally less costly to produce the results.
Algorithm training was done in batches of $128$ patches which was found to be suitable. The trained model had overall 743209 learnable parameters. Batch iterative training in the nolearn and lasagne functions was replaced with theano's LeNet 5~\cite{LBBH98} early stopping algorithm which showed further reduction in validation error relative to the train error.
\begin{figure}[!htb]
\begin{center}
\centerline{\includegraphics[width=\columnwidth]{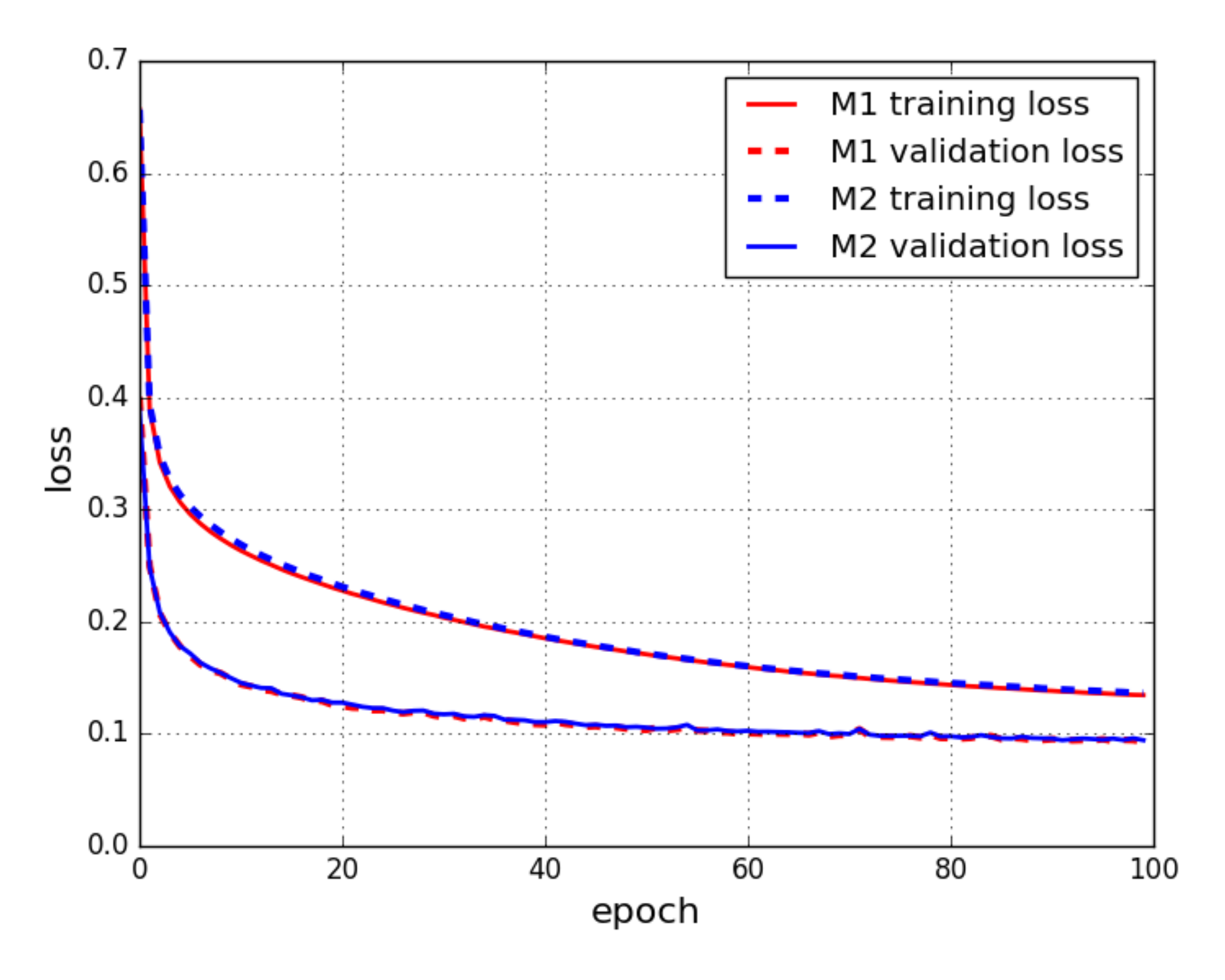}}
\caption{Results showing training and validation errors minimization progress plot for two model structures shown in fig. \ref{fig:architecture} with selected hyperparameters}
\label{fig:error}
\end{center}
\vskip -0.3in
\end{figure} \\
In the progress plot of fig.~\ref{fig:error} shown, the effect of our regularizers are to raise artificially the training error over the validation error in order to ensure accommodation of more training epoch. More training epochs are required for the lowered learning rates as pointed earlier while the learning rate was optimal to allow gradual convergence to the minimum achieved.
\begin{defn}(\textbf{Model structures}) \label{defn1}
Two different models using ${3 \times 3}$ receptive field were explored as shown in fig.~\ref{fig:architecture}.\\
\underline{Model 1}: (compressed decoder) Used $96$ feature maps at the unpooling and deconvolution layers. The model limits spreading of compressed information over many maps. However, it may suffer from discarding information due to capacity.\\
\underline{Model 2}: (uncompressed decoder) Used $128$ feature maps at the unpooling and deconvolution layers. The model has the capability of the capturing more information of the fully-connected layer with a potential problem of high information entropy.
\end{defn}
\begin{figure*}[!htb]
\begin{center}
 \includegraphics[width=1.0\textwidth,trim={0 0 10 0}]{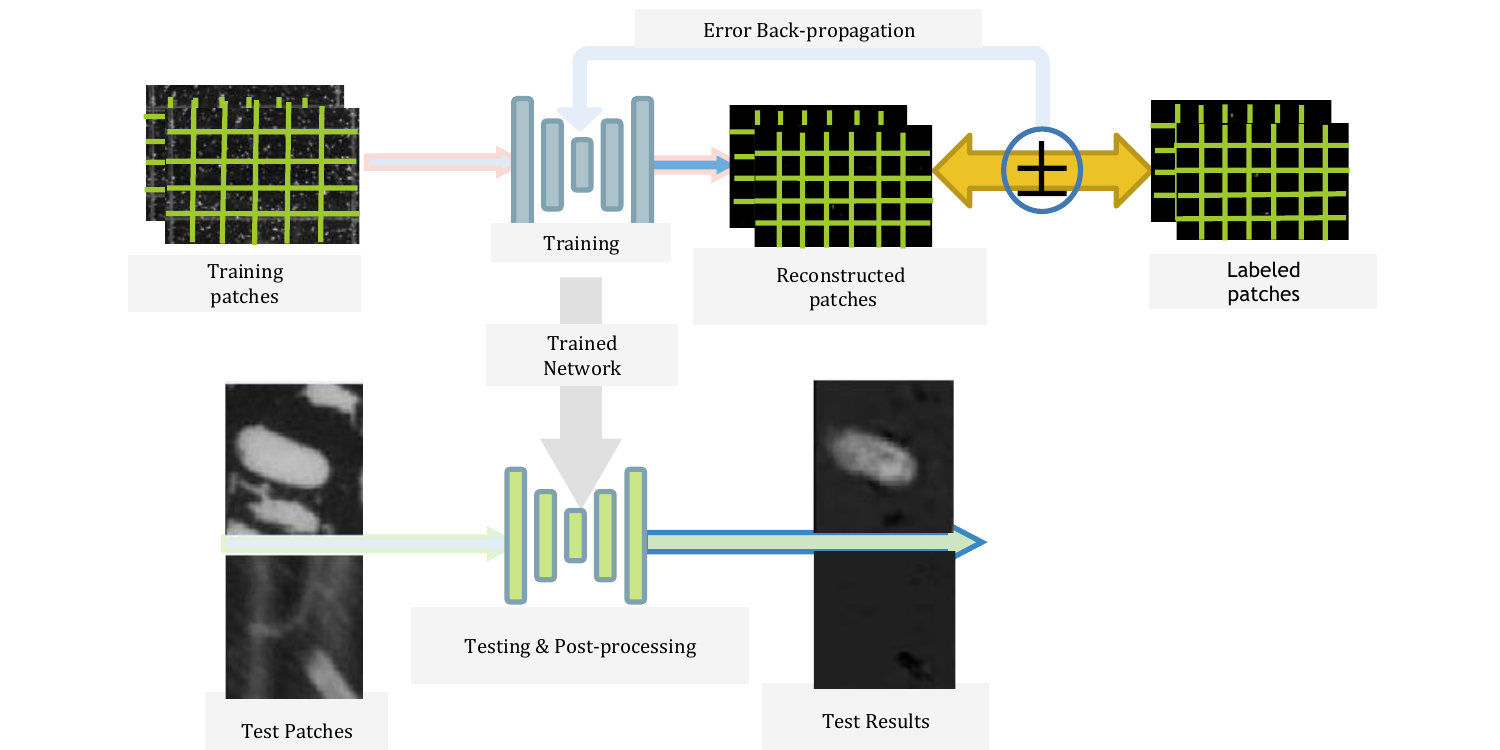}
\caption{Flow diagram of algorithm's implementation}
\label{fig:implement}
\end{center}
\vskip -0.3in
\end{figure*}
\textbf{Testing set}:
Testing process used $P = U \times V$, $(m \times n)$ = $(16 \times 16)$ patches at strides $(s_h, s_w)$
where $U$ and $V$ are the final vertical and horizontal number of patches respectively which are expressed as, 
\begin{equation} \label{equ:sh}
U = \frac{(M - m + s_h)}{s_h }
\end{equation}
\begin{equation} \label{equ:sw}
V = \frac{(N - n + s_w)}{s_w}
\end{equation}
\\
\textbf{Post-processing algorithm}:
In order to reduce false alarms especially when non-egg particles have high degree of similarity with the eggs, two forms of postprocessing around local neighborhoods in each frame were explored and described in Algorithm ~\ref{alg:post}.
\begin{algorithm}[!htb]
   \caption{Difference and Mean(ave)/ Maximum(max)} 
   \label{alg:post}
\begin{algorithmic}
   \STATE {\bfseries Input:} array = $(P \times m \times n)-$dimensional, patch size = $(m, n)$, $val$ = threshold value
   \STATE Initialize: $image = zeros(M \times N)$
   \FOR{$h = 1$ {\bfseries to} $U$ in Eqn~\ref{equ:sh}}
   \FOR{$w = 1$ {\bfseries to} $V$ in Eqn~\ref{equ:sw}}
   \STATE $x = h \times w$
   \IF {$(max(array^x) - min(array^x)) \leq val$}
   \STATE $array^{hw} = 0$
   \ENDIF
   \STATE $image^x_{m,n} = ave(array^x, image^x_{m,n})$
   \OR
   \STATE $image^x_{m,n}  = max(array^x, image^x_{m,n})$
   \ENDFOR
   \ENDFOR
\end{algorithmic}
\end{algorithm}\\
Therefore, a thresholded differencing was added to the non-maximum suppression scheme ~\cite{MGE11}.
The flow chart in fig.~\ref{fig:implement} shows the schematics of the overall implementation of the tool-chain including patching, convnet training and the post-processing steps.\\
\section{Results and discussions}
\label{sec:results}
In this section, performance of the convolutional selective autoencoder is presented and analyzed with respect to the SCN egg detection problem. The analysis is divided into two main parts: detection effectiveness (from an algorithmic perspective) and computation time and accuracy (from an application requirement perspective).
\begin{figure}[!htb]
\begin{center}
\centerline{\includegraphics[width=\columnwidth]{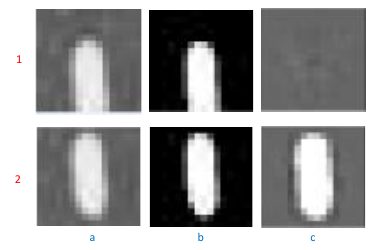}}
\caption{Plate 1a). uncentered patch, 1b). label for 1a, 1c). algorithm's result for 1a, 2a). centered patch, 2b). label for 2a and 2c). algorithm's result for 2a}
\label{fig:pbresult}
\end{center}
\vskip -0.3in
\end{figure}
Before discussing the algorithm's detection effectiveness, a justification of the pipeline's ability to reproduce the patch blocking training is shown in the fig.~\ref{fig:pbresult}.
\subsection{Detection effectiveness}  
\begin{figure*}[!htb]
\centering
 \includegraphics[width=1.0\textwidth,trim={0 0 10 0}]{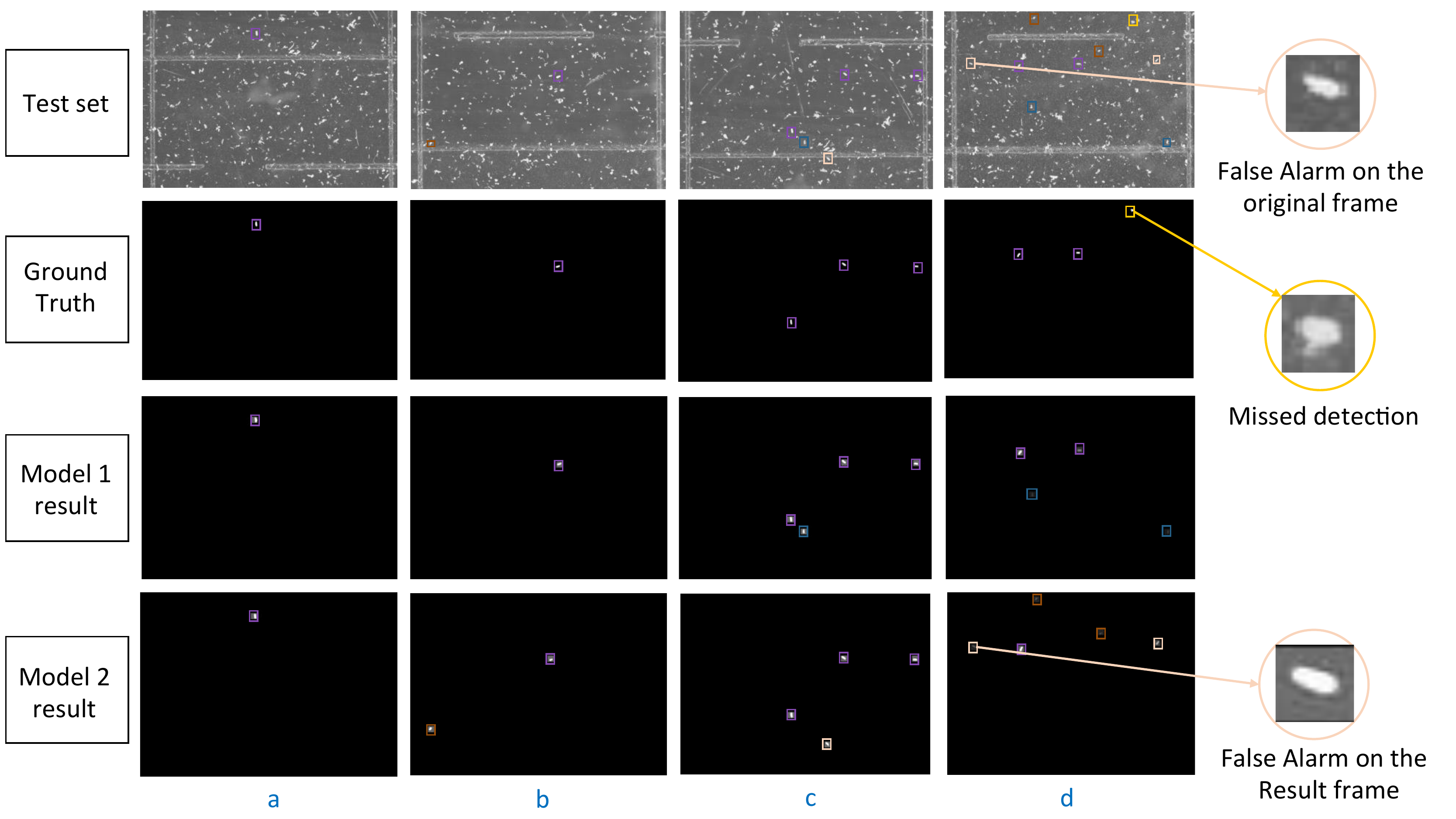}
\caption{Detection results with purple boxes indicating correctly labeled eggs, deep blue and light blue boxes for Model 1's clear and contentious (possibly human labeling error) false alarms respectively, deep orange and light orange boxes for Model 2's clear and contentious false alarms respectively and yellow boxes for missed detection. The gray scale threshold values of Models 1a).192, 1b).179, 1c).180, 1d).180, 2a).193, 2b).180, 2c).187 and 2d).173}
\label{fig:main_result}
\vskip -0.20in
\end{figure*}

Egg detection results obtained from the convnet-based tool-chain for $4$ randomly chosen testing sets are shown in fig~\ref{fig:main_result}.

Results shown on plates a and b of fig.~\ref{fig:main_result} where the algorithm captures the eggs (only) shows its effectiveness in suppressing the neighboring non-egg particles. In these cases, properties such as shape, pose, illumination of the non-egg particles are reasonably different from those of the eggs particles. This would especially be true in the local neighborhood of the eggs where any influence of highly similar non eggs would easily have influenced the result negatively.

On plate fig.~\ref{fig:main_result}c, both model types have one false alarm at different locations for an optimal post-processing threshold value. While the influence of local neighbor may not be so large, the possibility of a mislabeled non-egg particles by human labeler may provide a suitable reason for such an anomaly as shown in the result of Model 2. For example, one may argue that for the highlighted false alarm on plate fig.~\ref{fig:main_result}d, it as an egg indicating large possibility of human-labeling error. Therefore, the tool-chain can potentially help experts identify some of their probable defects in identifying the eggs as well as remove the bias in detection since human decisions are subject to changes of interpretations. Model 2 results generally seem to be show more "false alarms" possibly due to its enlarged feature maps. This is also supported by its usual low probability for actual labeled ground truth eggs detected by Model 1. The latter on the contrary usually detects eggs with high fidelity and no doubtful misses were recorded.

An envisaged defect of our proposed framework would be the non-detection of boundary situated eggs. However, an end correction scheme was added to the framework. This is a zero-padding type scheme to extend each test frame beyond its boundaries to ensure that those eggs with edges directly on the boundary of the frame are centered sometimes. In this study, we added padding with size same as that of a patch on all sides of an image frame. This ensures that the patching process will effectively cover an object on the image boundary, allowing the algorithm to enclose an egg completely with higher probability, at least once for a high resolution postprocessing.
\begin{figure}[!htb]
\begin{center}
\centerline{\includegraphics[width=\columnwidth]{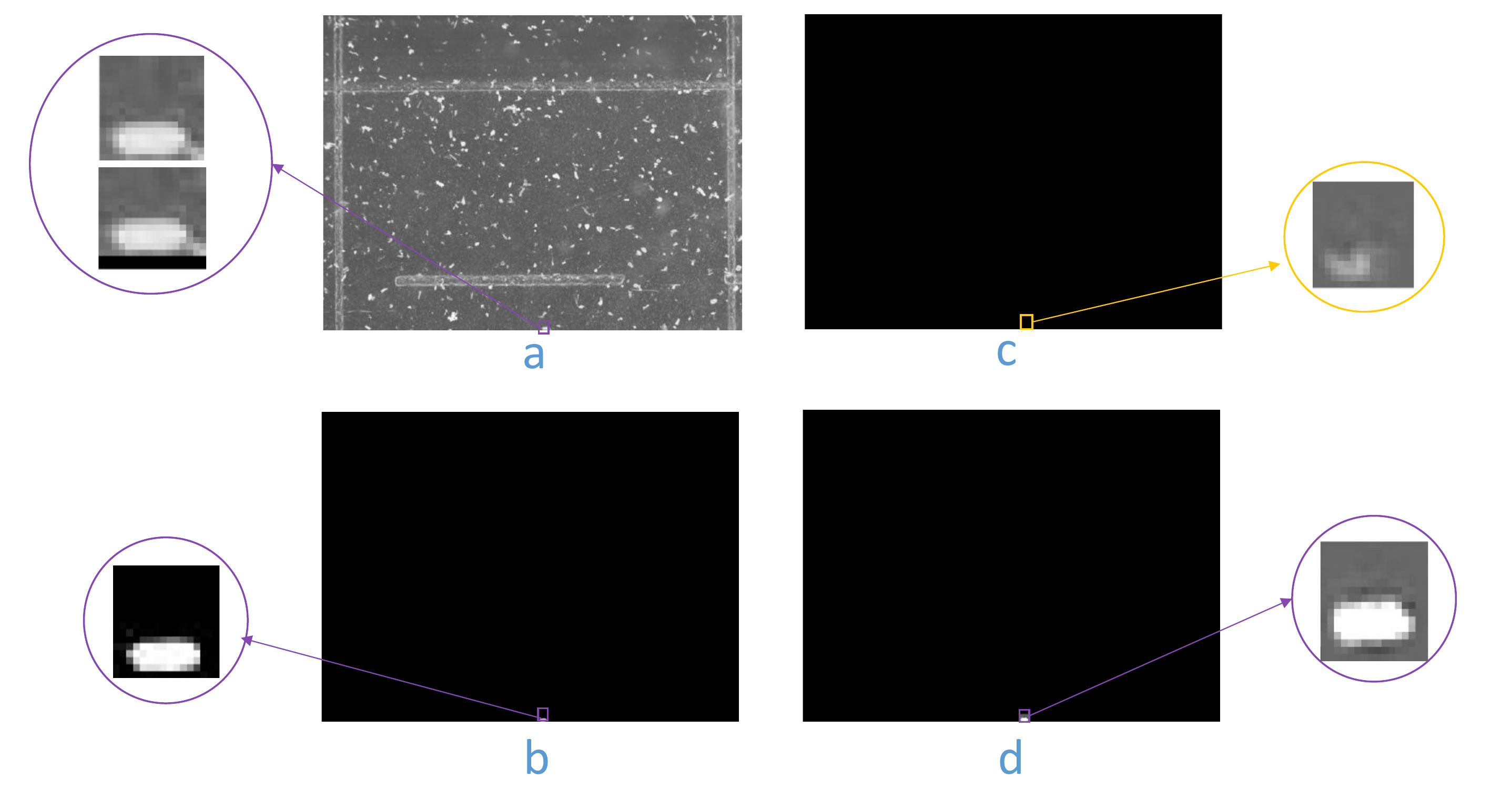}}
\caption{Boundary situated object (egg) scenario: (a) test frame, (b) ground truth with egg shown in a purple box, (c) missed detection (shown in yellow box) due to lack of boundary padding and (d) successful detection (shown in purple box) as a result of end correction}
\label{fig:boundary}
\end{center}
\vskip -0.5in
\end{figure}
Figure~\ref{fig:boundary} shows a situation with a boundary situated egg. A low activation of the plate c causes a missed detection of the boundary situated egg due to lack of boundary padding whereas end correction enables a successful detection as shown in plate d.

\subsection{Computation time and accuracy}
While training the convnet model with the large data sets as described earlier takes several hours aided by our GPU parallelizing capability, testing to identify the eggs from the frames can be a much faster high throughput operation. Testing patches were created with an adequate stride during in order to reduce the number of misses of bounding box of an egg. All the previous results were generated with strides ($s_h=2$, $s_w=2$). However, a ($s_h=4$, $s_w=4$) for instance would mean that there is higher chance of having not fully enclosed eggs. Hence, with lower patch stride, accuracy increases at a cost of increased complexity. A ($s_h=1$, $s_w=1$) at constant $M$, $m$, $N$ and $n$ should be the best possible. However, it would require large memory and computation time. Formally, the computation complexity for prediction due to the patching step can be described as: 
\begin{equation}
  \mathcal{O}(\frac{(M - m + s_w) \times (N - n + s_w)}{s_h \times s_w}) \approx \mathcal{O}(M\times N)
 \end{equation}
Table~\ref{comp-time} shows the detection times required for an image frame with different patch strides. In comparison, a well-trained expert plant-pathologist may take in the order of $5$ minutes to examine a frame.

 \begin{table}[!htb]
\caption{Detection time required for an image frame with different patch sizes}
\label{comp-time}
\begin{center}
\begin{small}
\begin{sc}
\begin{tabular}{lcccr}
\hline
\abovespace\belowspace
stride & P - \#of patches/frame   & detection time(sec)\\
\hline
\abovespace
$1 \times 1$ & $358821$  & $435$ \\
$2 \times 2$ & $77361$  & $77.5$\\
$4 \times 4$ & $19481$ &$18.05$\\
$8 \times 8$ & $4941$  & $5.6$\\
$16 \times 16$ & $1271$ &$1.8$\\
\hline
\end{tabular}
\end{sc}
\end{small}
\end{center}
\end{table}
In most test cases, the ($2 \times 2$) stride provides a good trade-off between computation time and accuracy. 

Typical performance metrics used for object detection tasks such as the accuracy and the confusion matrix may be inadequate due to the overwhelming presence of non-egg particles. 
Therefore, the following three performance metrics were formulated specifically for the current rare object detection problem: 
average detection accuracy, ADA
\begin{center}
= $\frac{\mbox{\# of eggs detected}}{ \mbox{actual \# of eggs}}$\\
\end{center}
average miss-to-egg ratio, AMER
\begin{center}
= $\frac{\mbox{ \# of false alarms per frame}}{ \mbox{actual \# of eggs per frame}}$\\
\end{center}
average non-eggs discarded, AND
\begin{center}
= $\frac{\mbox{\# of non-eggs discarded per frame}}{ \mbox{\# of non-eggs originally per frame}}$\\
\end{center}
Based on the metrics, the combined result for all test frames is shown in table~\ref{accuracy}. Note, the post-processing thresholds are chosen with a higher preference on detection accuracy compared to lowering false alarms. The rationale is that with a low missed detection probabbility, the resulting frames can be quickly examined by the experts to reject the false alarms (which is drastically low in number compared to the non-egg objects in the original frames) and still have a reliable count of eggs.
 \begin{table}[!htb]
\caption{Performance comparison based metrics defined for rare object (egg) detection}
\label{accuracy}
\begin{center}
\begin{small}
\begin{sc}
\begin{tabular}{lcccr}
\hline
\abovespace\belowspace
Model $\#$ & ada(\%)&amer(\%)&and(\%)\\
\hline
\abovespace
1& 94.33& 18.18& 99.77& \\
2& 83.17& 36.36& 99.30&\\
\hline
\end{tabular}
\end{sc}
\end{small}
\end{center}
\end{table}

It is clear that the models are very efficient in discarding most of the non-egg particles (metric AND). While the detection performance is also significantly high (metric ADA), false alarms are mostly caused due to objects with very similar characteristics as eggs (metric AND). However, many of the false alarms can be caused by human labeling errors and therefore, the tool-chain may require re-verification step by the experts to adaptively improve the model performance. Note that one of the assumptions of this selective autoencoder framework is that the patch size must be at least same or larger compared to the size of the largest SCN egg to be detected. 
\section{Summary, conclusions and future work}\label{sec:summary}
An end-to-end convolutional selective autoencoder approach is developed for a complex rare object detection problem. Hyperparameters and model structures for the convolutional network are meticulously explored for a critical plant science problem regarding automated detection of SCN eggs in microscopic images of soil samples.
The machine learning pipeline uses expert-labeled training examples (with the possibility of human-errors) and can serve as a decision support tool that has potential of saving enormous time of agricultural scientists in characterizing a significant disease affecting soybean yield in the United States.
From a machine learning perspective, a major issue is that a typical image frame in this application mostly contain other objects that are extremely similar to the objects of interest (SCN eggs). Therefore, hand-crafting features become a very difficult proposition and hence, deep learning becomes an appropriate choice. The following research areas are currently being pursued: (i) improvement of pre-processing the object patches via learning optimal transformation for a more efficient detection; (ii) adaptively fuse decision from multiple deep architectures ; (iii) exploring various unpooling strategies and interfacing classifiers at the fully connected layers  and (iv) learning to automatically count the rare objects within the automated pipeline.

\section*{Acknowledgements}
Authors gratefully acknowledge the support of NVIDIA Corporation with the donation of the GeForce GTX TITAN Black GPU used for this research.

\bibliography{arxivSCN}
\bibliographystyle{icml2016}

\end{document}